\pdfoutput=1

\documentclass{article}

\usepackage[final]{neurips_2023}

\usepackage{marvosym}
\usepackage{natbib}
\setcitestyle{numbers,square}

\usepackage[utf8]{inputenc} 
\usepackage[T1]{fontenc}    
\usepackage{hyperref}       
\usepackage{url}            
\usepackage{booktabs}       
\usepackage{amsfonts}       
\usepackage{nicefrac}       
\usepackage{microtype}      
\usepackage{xcolor}         

\usepackage{graphicx}
\usepackage{multirow}
\usepackage{arydshln}

\usepackage{bm}

\usepackage{caption}

\newcommand{\lxm}[1]{\textcolor[rgb]{0,0,0}{#1}}

\usepackage{xspace}

\makeatletter
\DeclareRobustCommand\onedot{\futurelet\@let@token\@onedot}
\def\@onedot{\ifx\@let@token.\else.\null\fi\xspace}

\def\eg{\emph{e.g}\onedot} 
\def\ie{\emph{i.e}\onedot}

\def\etal{\emph{et al}\onedot}
\makeatother

\title{Ref-Diff: Zero-shot Referring Image Segmentation with Generative Models}

\author{%
  Minheng Ni$^1$\ \ Yabo Zhang$^1$\ \ Kailai Feng$^1$\ \ Xiaoming Li$^1$\ \ Yiwen Guo$^3$\ \ Wangmeng Zuo$^{1,2}$\textsuperscript{~\Letter} \\
$^1$Harbin Institute of Technology\quad $^2$Pengcheng Laboratory\quad $^3$Independent Researcher\\
  \texttt{mhni@stu.hit.edu.cn}\quad \texttt{hitzhangyabo2017@gmail.com}\quad \texttt{klfeng@stu.hit.edu.cn}\\
  \texttt{csxmli@gmail.com}\quad \texttt{guoyiwen89@gmail.com}\quad \texttt{wmzuo@hit.edu.cn}\\
}

\begin{document}

\maketitle

\begin{abstract}
Zero-shot referring image segmentation is a challenging task because it aims to find an instance segmentation mask based on the given referring descriptions, without training on this type of paired data. Current zero-shot methods mainly focus on using pre-trained discriminative models (\eg, CLIP). However, we have observed that generative models (\eg, \textsc{Stable Diffusion}) have potentially understood the relationships between various visual elements and text descriptions, which are rarely investigated in this task. In this work, we introduce a novel Referring Diffusional segmentor (\textsc{Ref-Diff}) for this task, which leverages the fine-grained multi-modal information from generative models. We demonstrate that without a proposal generator, a generative model alone can achieve comparable performance to existing SOTA weakly-supervised models. When we combine both generative and discriminative models, our \textsc{Ref-Diff} outperforms these competing methods by a significant margin. This indicates that generative models are also beneficial for this task and can complement discriminative models for better referring segmentation. Our code is publicly available at \href{https://github.com/kodenii/Ref-Diff}{https://github.com/kodenii/Ref-Diff}.
\end{abstract}

\section{Introduction}

Referring Image Segmentation (RIS) aims to identify the referring instance region that is semantically consistent with the given text description. Different from semantic segmentation, this task often requires distinguishing instances of the same class, \eg, the tallest boy among these four children. Annotation of precise pairs (\ie, image, text description, and ground-truth instance mask) is costly and time-consuming. A recent weakly-supervised RIS approach~\cite{strudel_weakly-supervised_2022} endeavors to alleviate the annotation difficulties, but it still needs specific pairs of images and referring texts for training. Conversely, a zero-shot solution is more valuable but may exacerbate the challenges further. On the one hand, it is training-free, and no referring annotation is required. On the other hand, it needs a deeper comprehension of the relationship between text and the visual elements in the images.

Recent multi-modal pre-training models have shown impressive capabilities in vision and language understanding. As one of the most representative discriminative models among them, CLIP~\cite{radford_learning_2021} (which explicitly learns the global similarity between image and text through contrastive learning) has demonstrated significant improvements to various tasks, including object detection~\cite{zhong_regionclip_2021}, image retrieval~\cite{radford_learning_2021}, and semantic segmentation~\cite{dong_maskclip_2023}. 
However, directly applying a model like CLIP to zero-shot RIS is impractical, as it is trained to capture the global similarity of text and images, which cannot well learn the specific visual elements relating to a referring text. To address this, Yu~\etal~\cite{yu2023zero} propose a global and local CLIP to bridge the gap between discriminative models and pixel-level dense prediction. Nevertheless, we observe that discriminative models themselves struggle to localize visual elements accurately.
In recent years, generative models such as \textsc{Stable Diffusion} \cite{rombach_high-resolution_2022}, DALL-E 2 \cite{ramesh_hierarchical_2022}, and \textsc{Imagen} \cite{saharia2022photorealistic} have also attracted great attention due to their ability in generating the realistic or imaginative images. The semantic alignment in generated images demonstrates that these generative models have implicitly captured the relationships between various visual elements and texts. However, unlike discriminative models, they are rarely exploited in zero-shot referring image segmentation tasks.

In this work, we attempt to investigate whether the generative models can benefit the zero-shot RIS task. To this end, we propose a novel Referring Diffusional segmentor (\textsc{Ref-Diff}). It leverages fine-grained multi-modal information from generative models to exploit the relationship between referring expressions and different visual elements in the image. Previous works usually adopt CLIP to rank the proposals from an offline proposal generator~\cite{kirillov_segment_2023}. In contrast, our \textsc{Ref-Diff} can inherently provide these instance proposals using the generative models. This indicates that our \textsc{Ref-Diff} does not necessarily depend on third-party proposal generators.

\lxm{Experiments on three datasets show that without the use of an offline proposal generator, only the generative model in our \textsc{Ref-Diff} achieves comparable performance against the SOTA weakly-supervised methods. Additionally, when incorporating an offline proposal generator and discriminative model, our Ref-Diff significantly outperforms the competing methods. Both quantitative and qualitative analyses demonstrate that the generative model is beneficial for this task, and the combination with discriminative models can lead to better referring segmentation results.}

\lxm{
The main contributions can be summarized as follows:
\begin{itemize}
    \item We demonstrate that the generative models can be leveraged to improve the zero-shot RIS task by exploiting the implicit relationships between visual elements and text descriptions.
    \item We show that the generative model can intrinsically perform proposal generation, thereby making our \textsc{Ref-Diff} independent of third-party proposal generators.
    \item We propose a feasible manner to combine the generative and discriminative models for the zero-shot RIS task, which complement each other and achieve better referring segmentation.
\end{itemize}
}
\section{Related Work}

\paragraph{Zero-shot Referring Segmentation.}

\lxm{
Referring image segmentation is one of the most fundamental and challenging tasks, as it involves a fine-grained understanding of both vision and language.
Following a fully-supervised formulation, previous works~\cite{xu2023meta,yan2023universal,yang2023semantics,liu2023polyformer,ding2022vlt,yang2022lavt,zhao2023unleashing} require labor-intensive training annotations, \ie, referring expressions and pixel-level masks.
However, due to the absence of large-scale training annotations, these methods are often limited in their scalability and out-of-domain samples~\cite{strudel_weakly-supervised_2022}.
With the remarkable progress of discriminative vision-language pre-training~\cite{radford_learning_2021}, recent works~\cite{strudel_weakly-supervised_2022,yu2023zero} explore their open-vocabulary recognition in weakly or zero-shot referring segmentation.
Despite their considerable performance, the pre-trained discriminative models that learn the global similarity of text and image have inherent limitations in deeply understanding the object delineation or the fine-grained relationships between visual elements and text description.
In contrast, our \textsc{Ref-Diff} utilizes the fine-grained understanding through generative models and thereby obtains more accurate predictions.
}

\paragraph{Visual Generative Models for Non-Synthesized Tasks.}
\lxm{Large-scale text-to-image generative models~\cite{rombach_high-resolution_2022,ramesh_hierarchical_2022,saharia2022photorealistic} have achieved tremendous progress in imaginary generation and creative applications~\cite{wei2023elite,ruiz2022dreambooth,gal2022image,controlnet,mou2023t2i}.
Apart from generation-related tasks, these generative models also demonstrate preeminent capabilities in fine-grained image understanding, \eg, semantic segmentation~\cite{zhao2023unleashing,wu2023diffumask,asiedu2022decoder,baranchuk2021label,amit2021segdiff}, object detection~\cite{cheng2023parallel,chen2022diffusiondet}, dense prediction~\cite{ji2023ddp,saxena2023monocular}, and classification~\cite{li2023your,clark2023text}.
Therefore, most existing research focuses on transfer learning or constructing synthetic data for specific tasks. Although a few studies have explored the application of using generative models in zero-shot image classification, the performance of generative models in zero-shot referring segmentation has not been well investigated and deserves to attract attention.
}
\section{Methodology}

\begin{figure*}
	\centering
	\includegraphics[width=14cm]{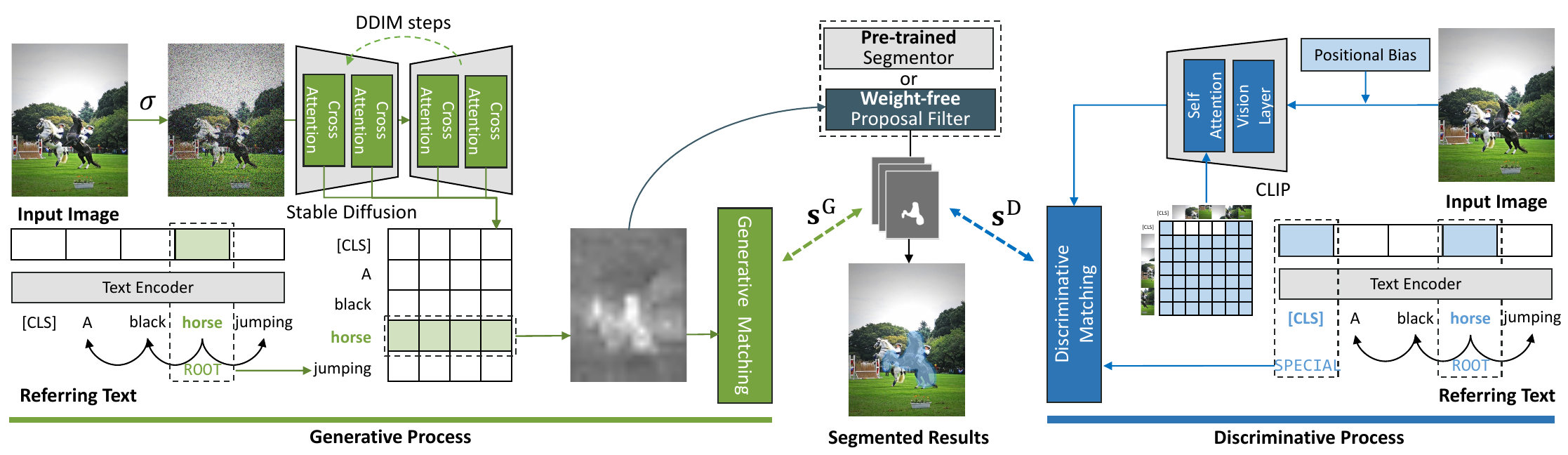}
    \caption{\lxm{Overview of our \textsc{Ref-Diff}. Our proposed Generative Process (left) generates a correlation matrix between the referring text and the input image. This matrix serves as an alternative weight-free proposal generator and generative segmentation candidates $\mathbf{s}^{\mathrm{G}}$. The Discriminative Process (right) is alternatively integrated into our framework and generates the discriminative candidates $\mathbf{s}^{\mathrm{D}}$. The final referring segmentation result is obtained either from the generative candidates or a combination of both generative and discriminative candidates.}}
	\label{fig:framework}
\end{figure*}

\subsection{Problem Formulation and Inference Pipeline}

Given an image $\mathbf{x}\in\mathbb{R}^{W\times H\times C}$ and a referring text $T$, Referring Segmentation aims to output a segmenting mask $\mathbf{m}\in\{0,1\}^{W\times H}$ indicating the referring regions of the text $T$ in image $\mathbf{x}$, where $W$, $H$, and $C$ represent the width, height, and channel of the image, respectively. In the Zero-shot Referring Segmentation settings, the model cannot access any training data of Referring Segmentation, including images, referring texts, and instance mask annotations.

Our proposed framework is depicted  in Figure. \ref{fig:framework}. Given an image and the referring text, our \textsc{Ref-Diff} generates a correlation matrix using the Generative Process, which can be used as 1) an alternative weight-free proposal generator and 2) a set of referring segmentation candidates. Optionally, our \textsc{Ref-Diff} can integrate the discriminative model with our proposed generative model within a unified framework. The final similarity for each mask proposal is obtained as follows:
\begin{equation}
    \mathbf{s}_i = \alpha\mathbf{s}^{\mathrm{G}}_i + (1 - \alpha)\mathbf{s}^{\mathrm{D}}_i\,,
\end{equation}
where $\alpha$ is a hyper-parameter. $\mathbf{s}^{\mathrm{G}}_i$ and $\mathbf{s}^{\mathrm{D}}_i$ are the generative and discriminative scores between the referring text and $i$\textit{-}th proposal, respectively. When $\alpha$ is set to $1$, only the generative model is adopted for this task. 
The final referring segmentation result is determined by selecting the proposal with the highest similarity score:
\begin{equation}
    \hat{\mathbf{m}} = \mathop{\arg\max}\limits_{\mathcal{M}_{i}}\mathbf{s}_i.
\end{equation}
In the following sections, we provide a detailed introduction to each module of our framework.

\subsection{Generative Process}

\textsc{Stable Diffusion}~\cite{rombach_high-resolution_2022} is an effective generative model that consists of a series of inverse diffusion steps to gradually transform random Gaussian noise into an image. 
Therefore, it cannot directly operate on a real image to obtain its latent representations. 
Fortunately, as the diffusion process is computable, we can take the real image as one intermediate state generated by a diffusion model and run it backward to any step of the generation. 
In this work, we add a specific amount of Gaussian noise to obtain $\mathbf{x}_t$ and then continue this process without compromising information:
\begin{equation}
    \mathbf{x_t} = \sigma_t(\mathbf{x}),
\end{equation}
where $\sigma_t$ is the function to obtain the noised image in step $t$ and $t$ is a hyper-parameter.

During the inverse diffusion process, let $\Psi_{\mathrm{lan}}$ and $\Psi_{\mathrm{vis}}$ be the text and image encoder of the generative model, respectively. 
In the generative model, the referring text $T$ is encoded into text features using $\mathbf{K} = \Psi_{\mathrm{lan}}(T) \in \mathbb{R}^{l\times d}$, where $l$ is the token number and $d$ is the dimension size for latent projection. Similarly, for the $i$\textit{-}th step, the visual image $\mathbf{x_i}$ is projected to image features using $\mathbf{Q} = \Psi_{\mathrm{vis}}(\mathbf{x_i}) \in \mathbb{R}^{w\times h\times d}$. Here, $w$ and $h$ are the width and height of encoded image features.
The cross-attention between the text and image features can be formulated as:
\begin{equation}
    \mathbf{a} = \mathrm{Softmax}(\frac{\mathbf{Q}\mathbf{K}^{\top}}{\sqrt{d}})\,,
\end{equation}
where $\mathbf{a} \in \mathbb{R}^{w\times h \times
l \times N}$, and $N$ is the number of attention heads. Following~\cite{wu2023diffumask}, we obtain the overall cross-attentions by averaging the value of each attention head to $\bar\mathbf{a} \in \mathbb{R}^{w\times h \times l}$. The cross-attention matrix $\bar\mathbf{a}$ represents the correlation detected between each token in the referring text and each region feature in the image. In general, a higher value in $\bar \mathbf{a}$ indicates a better correlation between the token and region features, which can be used to locate the related referring regions.

In the inverse diffusion process, the generative model captures the overall semantics of the language condition. However, the corresponding attention region for each token is not necessarily the same as they have different semantic representations (see Sec.~\ref{sec:attn}). Without loss of generality, a referring text $T$ is a sentence that describes the characteristics of a specific instance. To obtain the preferred token from the whole text description, we use syntax analysis to obtain its root token (\ie, the \texttt{ROOT} element in the syntax tree). Generally, the \texttt{ROOT} token in the latent space captures the contextual correlations (\eg, the token 'horse' in Figure~\ref{fig:framework} contains contextural representations from 'black' and 'jumping'). Then, the attention region projected for this root token has a higher probability of being the referring region. Let $k$ be the index of the root token, and let $\bar\mathbf{a}_k \in \mathbb{R}^{w\times h}$ denote the cross-attention matrix of the root token.
We normalize and resize this cross-attention matrix by:
\begin{equation}
    \mathbf{c} = \phi_{w\times h\to W\times H}\left({\frac{\bar\mathbf{a}_k - \mathrm{min}(\bar\mathbf{a}_k)}{ \mathrm{max}(\bar\mathbf{a}_k) - \mathrm{min}(\bar\mathbf{a}_k) + \epsilon}}\right)\,,
\end{equation}
where $\epsilon$ is a small constant value. Here, $\phi_{w\times h\to W\times H}$ is a bi-linear interpolation function used to resize the attention map to the same resolution as the given image.

\subsection{Discriminative Process}\label{sec:dis}

During the image encoding process by the discriminative model CLIP, the spatial position is inevitably attenuated. We observe that referring text descriptions usually contain explicit direction clues (\eg, left, right, top, and bottom), which are valuable but have been ignored in previous works.
To emphasize such types of positional information, we propose a positional bias to explicitly encode the image with the given direction clues. This is achieved through element-wise multiplication:
\begin{equation}
    \mathbf{x}' = \mathbf{x} \odot \mathbf{P}\,,
\end{equation}
where $\mathbf{P}\in \mathbb{R}^{W \times H \times C}$ is a positional bias matrix. Specifically, if the text, after syntactic analysis, contains explicit direction clues, $\mathbf{P}$ will be a soft mask with values ranging from $1$ to $0$ along the given direction axis. Lower values indicate regions that should receive less attention. Conversely, if no direction clue is detected, $\mathbf{P}$ will be a matrix filled with $1$.

Finally, the ultimate representation $\mathbf{v}_i \in \mathbb{R}^{d}$ for each proposal $\mathcal{M}_i$ in discriminative process is:
\begin{equation}
    \mathbf{v}_i = \beta f_{\mathcal{M}_i}(\mathbf{x} \odot \mathbf{P})\ + (1-\beta)f(\mathbf{x} \odot \mathcal{M}_i),
\end{equation}
where $f$ and $f_{\mathcal{M}_i}$ is the vanilla CLIP image encoder and CLIP image encoder with modified self-attention based on mask proposal $\mathcal{M}_i$. Since the discriminative model (\ie, CLIP) is expected to encode the instance within each proposal region $\mathcal{M}_i$ while disregarding other regions for reducing disturbances. To achieve this, we assign a weight of $0$ to the attention values between the \texttt{[CLS]} token and the patch tokens outside the current proposal $\mathcal{M}_i$.
In this work, we utilize the output of the penultimate layer as the final representation, which is motivated by the observation that the representation in the last layer tends to encompass the entire image rather than focus on the corresponding proposal region.

\subsection{Proposals Extracting and Matching}

\textbf{Weight-free Proposal Filter.} Since the generative models inherently encode instance representations, we can derive proposals from their cross-attention matrix $\mathbf{c}$. In this work, we introduce a weight-free proposal filter to generate a series of mask proposals. This is formulated as:
\begin{equation}
    \mathcal{M} = \left\{\psi(\mathbf{c}\geq\mu)|\mu\in\left\{5\%, 10\%, ..., 95\%\right\}\right\}\,,
    \label{eq:proposal}
\end{equation}
where $\psi$ is a binarization function with a given predefined threshold value $\mu$. Different from other works~\cite{yu2023zero} which rely on external proposal generators and CLIP filters, the generative models in this work can efficiently and effectively produce the expected proposals. This approach offers a streamlined and integrated solution for obtaining high-quality proposals without additional tools.

\lxm{
\textbf{Pre-trained Segmentor.} If a reliable segmentor is available, we can also obtain proposals from it in a flexible manner. By leveraging the capability of generative and discriminative models for semantic understanding, we can refine the proposals and prioritize those that align closely with the given referring description. This combined approach of using a segmentor for initial proposal generation ensures that the resulting proposals are coherent and better aligned with the given referring expression.
}

\textbf{Generative Matching.} After obtaining proposals either from the weight-free proposal filter or the pre-trained segmentor, the next step is to find the most similar proposal based on the cross-attention matrix. In this work, we quantify the similarity between the given referring text and all the proposals by measuring the distance on cross-attention matrix $\mathbf{c}$ and $\mathcal{M}_i$:
\begin{equation}
    \mathbf{s}^{\mathrm{G}}_i = \frac{\left|\mathbf{c}\odot\mathcal{M}_i\right|}{\left|\mathcal{M}_i\right|} - \frac{\left|\mathbf{c}\odot(1 - \mathcal{M}_i)\right|}{\left|1-\mathcal{M}_i\right|}.
\end{equation}

\textbf{Discriminative Matching.}  Given a referring text, we obtain the mean representation $\mathbf{r}\in \mathbb{R}^{d}$ of the global text and the local subject token using the CLIP text encoder, which serves as the features of the referring text. To find out the most probable proposal from the perspective of the discriminative model, we calculate the similarity between the features of referring text $\mathbf{r}$ and visual representation $\mathbf{v}_i$ of each mask proposal $\mathcal{M}_i$, which is defined as:
\begin{equation}
    \label{eqn:dis}
    \mathbf{s}^{\mathrm{D}}_i = \mathbf{v}_i\mathbf{r}^{\top}\,.
\end{equation}
This similarity allows us to identify the proposal that best aligns with the referring text. Higher similarity scores indicate a stronger correspondence between the proposal and the text, indicating a higher likelihood of being the correct segmentation result.

\section{Experiments}

\subsection{Implemention Details}
\begin{figure*}
	\centering
	\includegraphics[width=14cm]{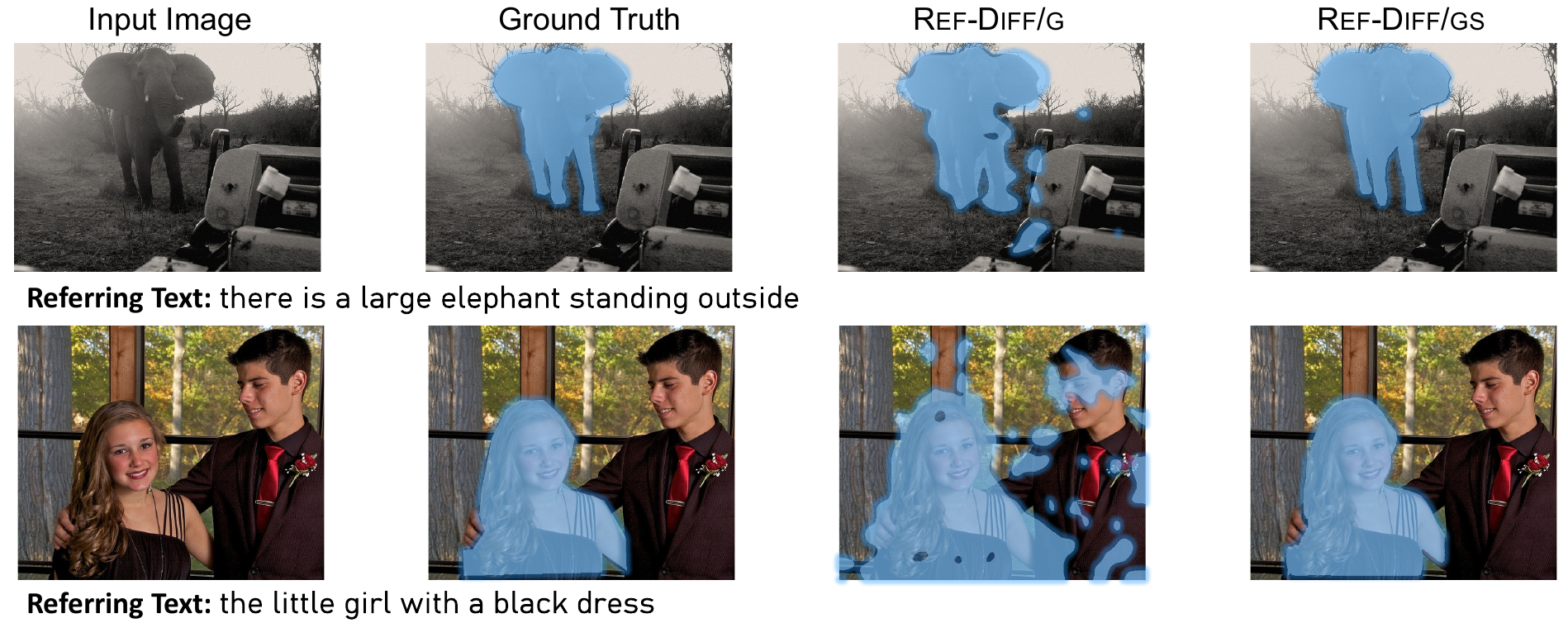}
	\caption{\textbf{Effectiveness of generative model in segmentation capability.} \textsc{Ref-Diff/g} is capable of segmenting the right content even without the assistance of the pre-trained segmentor and CLIP. Combing with the pre-trained segmentor, \textsc{Ref-Diff/gs} achieves precise segmentation of the correct regions.}
	\label{fig:nop}
\end{figure*}
Our \textsc{Ref-Diff} is a zero-shot solution, so we only need an inference process, without any training images and annotations. All experiments are conducted on a Tesla A100 GPU. We use the pre-trained \textsc{Stable Diffusion}~\cite{rombach_high-resolution_2022} (V1.5) as our generative model. All test images are resized and padded to the resolution of $1024\times 1024$. Since existing works mainly focus on using pre-trained segmentor and discriminative model~\cite{radford_learning_2021}, for a fair comparison, we select SAM~\cite{kirillov_segment_2023} as the segmentor and CLIP of ViT-B/16 as the discriminative model. We set $t$, $\alpha$, and $\beta$ to $2$, $0.1$, and $0.3$, respectively.

\subsection{Experimental Setup}

Following~\cite{yu2023zero,strudel_weakly-supervised_2022}, we adopt mIoU and oIoU as the evaluation metrics and apply them to three widely-used benchmarks, including RefCOCO~\cite{nagaraja_modeling_2016}, RefCOCO+~\cite{nagaraja_modeling_2016}, and RefCOCOg~\cite{mao_generation_2016}. 
For fairly comparing with existing works, we conduct the experiments under two settings: a) Zero-shot RIS using a pre-trained segmentor and CLIP; b) Zero-shot RIS without a pre-trained segmentor and CLIP. The latter setting allows us to analyze the effectiveness of the generative model in this task.

For setting a), we select five competing baselines.
1) A weakly-supervised method TSEG~\cite{strudel_weakly-supervised_2022}. It is not open-sourced and only provides the validation results of mIoU. So we did not report its results on other settings and test sets.
2)$\sim$4) Three zero-shot baselines from~\cite{yu2023zero}, including \textsc{Region Token}, \textsc{Cropping}, and \textsc{Global-Local CLIP}.
5) A zero-shot baseline SAM-CLIP proposed by us. Note that Yu~\etal~\cite{yu2023zero} adopt \textsc{FreeSOLO}~\cite{wang_freesolo_2022} as their segmentor. However, considering the remarkable performance of SAM~\cite{kirillov_segment_2023} in segmentation, we propose SAM-CLIP as a new discriminative baseline.
Specifically, we first use SAM to extract all candidate proposals from the image and then leverage CLIP to identify the most relevant proposal using Eqn.~\ref{eqn:dis}. In this setting, our \textsc{Ref-Diff} combines both generative and discriminative models, and uses the same proposal segmentor as other methods.

\begin{table*}[ht!]
\footnotesize
\centering
\setlength{\tabcolsep}{4pt}
\caption{\textbf{The oIoU comparison on settings of using a pre-trained segmentor and CLIP.} 
The improvement is statistically significant with $p < 0.01$ under $t$-test.}
\begin{tabular}{lccccccccccc}
\hline
 \multirow{2}{*}{\textbf{Methods}}           & \multicolumn{3}{c}{\textbf{RefCOCO}}                     & & \multicolumn{3}{c}{\textbf{RefCOCO+}}                    & & \multicolumn{3}{c}{\textbf{RefCOCOg}}                     \\ \cline{2-4} \cline{6-8} \cline{10-12} 
& val            & test A         & test B         & & val            &  test A         & test B         & & val(U)            & test(U)        & val(G)         \\ \hline
\multicolumn{10}{l}{\textit{Weakly-supervised Method}} \\
TSEG~\cite{strudel_weakly-supervised_2022}                                                                & - & - & - & & - & - & - & & - & - & - \\
\hline
\multicolumn{10}{l}{\textit{Zero-shot Method}} \\
\textsc{Region Token}~\cite{yu2023zero}                                                                  & 21.71          & 20.31          & 22.63          & & 22.61          & 20.91          & 23.46          & & 25.52          & 25.38          & 25.29          \\
\textsc{Cropping}~\cite{yu2023zero}                                                                  & 22.73          & 21.11          & 23.08          & & 24.09          & 22.42          & 23.93          & & 28.69          & 27.51          & 27.70          \\
\textsc{Global-Local CLIP}~\cite{yu2023zero}                                                                  & 24.88 & 23.61 & 24.66 & & 26.16 & 24.90 & 25.83 & & 31.11 & 30.96 & 30.69 \\
\textsc{SAM-CLIP}                                                                 & 25.23 & 25.86 & 24.75 & & 25.64 & 27.76 & 26.06 & & 33.75 & 34.80 & 33.65 \\
 \textbf{\textsc{Ref-Diff}}                                                                 & \textbf{35.16} & \textbf{37.44} & \textbf{34.50} & & \textbf{35.56} & \textbf{38.66} & \textbf{31.40} & & \textbf{38.62} & \textbf{37.50} & \textbf{37.82} \\
\hline
\end{tabular}
\vspace{-0.2in}
\label{tab:total_oiou}
\end{table*}

\begin{table*}[ht!]
\footnotesize
\centering
\setlength{\tabcolsep}{4pt}
\caption{\textbf{The mIoU comparison on settings of using a pre-trained segmentor and CLIP.} 
The improvement is statistically significant with $p < 0.01$ under $t$-test.}
\begin{tabular}{lccccccccccc}
\hline
 \multirow{2}{*}{\textbf{Methods}}           & \multicolumn{3}{c}{\textbf{RefCOCO}}                     & & \multicolumn{3}{c}{\textbf{RefCOCO+}}                    & & \multicolumn{3}{c}{\textbf{RefCOCOg}}                     \\ \cline{2-4} \cline{6-8} \cline{10-12} 
& val            & test A         & test B         & & val            &  test A         & test B         & & val(U)            & test(U)        & val(G)         \\ \hline
\multicolumn{10}{l}{\textit{Weakly-supervised Method}} \\
TSEG~\cite{strudel_weakly-supervised_2022}                                                        & 25.95          & -              & -              & & 22.62          & -              & -              & & 23.41          & -              & -              \\ 
\hline
\multicolumn{10}{l}{\textit{Zero-shot Method}} \\
\textsc{Region token}~\cite{yu2023zero}                                                                  & 23.43          & 22.07          & 24.62          & & 24.51          & 22.64          & 25.37          & & 27.57          & 27.34          & 27.69          \\
\textsc{Cropping}~\cite{yu2023zero}                                                                  & 24.83          & 22.58          & 25.72          & & 26.33          & 24.06          & 26.46          & & 31.88          & 30.94          & 31.06          \\ 
\textsc{Global-Local CLIP}~\cite{yu2023zero}                                                                  & 26.20          & 24.94          & 26.56 & & 27.80          & 25.64          & 27.84          & &33.52 & 33.67 & 33.61 \\
\textsc{SAM-CLIP}                                                                 & 26.33 & 25.82 & 26.40 & & 25.70 & 28.02 & 26.84 & & 38.75 & 38.91 & 38.27 \\
 \textbf{\textsc{Ref-Diff}}                                                                & \textbf{37.21} & \textbf{38.40} & \textbf{37.19} & & \textbf{37.29} & \textbf{40.51} & \textbf{33.01} & & \textbf{44.02} & \textbf{44.51} & \textbf{44.26} \\
 \hline
\end{tabular}
\label{tab:total_miou}
\end{table*}

The mIoU and oIoU comparisons are shown in Tables~\ref{tab:total_oiou} and~\ref{tab:total_miou}, respectively. We can observe that \textsc{Ref-Diff} exhibits significantly superior performance compared to the competing methods and baselines. Benefiting from the combination of both generative and discriminative models, our \textsc{Ref-Diff} achieves an improvement of approximately $10$ mIoU on RefCOCO, RefCOCO+, and RefCOCOg datasets.
From the comparison between \textsc{SAM-CLIP} and other methods, only a slight improvement is observed. We analyze that segmentation from SAM is highly detailed (\eg, it may split a single object into multiple small parts), which easily increases the possibility of CLIP filtering out erroneous proposals by only using the discriminative model CLIP. These types of erroneous proposals have almost no overlap with the correct solution, resulting in only marginal improvement. By incorporating the generative model, the filtering of erroneous proposals is further mitigated, contributing to our superior performance. This also demonstrates that our improvement over SAM-CLIP stems from a deeper understanding facilitated by both our generative and discriminative models.

Furthermore, we conduct additional experiments on PhraseCut dataset in Table~\ref{tab:phr}. We can see that our \textsc{Ref-Diff} also outperforms the competing method \textsc{Global-Local} by a large margin on OIoU. This further validates the great generalization of our \textsc{Ref-Diff} on other datasets.

\begin{table*}[ht!]
\footnotesize
\centering
\caption{\textbf{Results on PraseCut.} Ref-Diff outperforms the prior model with a significant improvement.}
\begin{tabular}{lcc}
\hline
\textbf{Methods}            & \textbf{oIoU}                     & \textbf{mIoU}                   \\
\hline
\textsc{Global-local} & 23.64 & - \\
\textsc{\textbf{Ref-Diff}} & \textbf{29.42} & \textbf{41.75} \\
 \hline
\end{tabular}
\label{tab:phr}
\end{table*}

\subsection{Ablation Study}
We conduct an ablation study in Table~\ref{tab:abl_oiou} and Table~\ref{tab:abl_miou}, which show the effectiveness of different components. Notably, our \textsc{Ref-Diff/g} achieves comparable performance to the weakly supervised model TSEG~\cite{strudel_weakly-supervised_2022} on \textsc{RefCOCO+}, and outperforms it on \textsc{RefCOCOg} when using the generative model alone. When combined with the segmentor, \textsc{Ref-Diff/gs} consistently exhibits superior performance across all test sets to the weakly supervised model.
These observations collectively show that the generative model can not only perform proposal generation but also benefit the RIS task. 

\begin{table*}[ht!]
\footnotesize
\centering
\setlength{\tabcolsep}{5pt}
\caption{\textbf{The ablation study on oIoU.}}

\begin{tabular}{lcccccc}
\hline
\textbf{Methods}           & \textbf{Segmentor} & \textbf{Generative} & \textbf{Discriminative} & \textbf{RefCOCO}                     & \textbf{RefCOCO+}                   & \textbf{RefCOCOg} \\
\hline
\textsc{Ref-Diff/g} & & \checkmark &  & 16.18 & 17.23 & 20.39 \\
\textsc{Ref-Diff/gs} & \checkmark & \checkmark&  & 26.04 & 26.68 & 26.84 \\
\textsc{Ref-Diff/ds} & \checkmark & & \checkmark & 33.64 & 34.43 & 34.83 \\
\textsc{Ref-Diff} & \checkmark & \checkmark & \checkmark & \textbf{35.16} & \textbf{35.56} & \textbf{38.62} \\
 \hline
\end{tabular}
\label{tab:abl_oiou}
\end{table*}

\begin{table*}[ht!]
\footnotesize
\centering
\setlength{\tabcolsep}{5pt}
\caption{\textbf{The ablation study on mIoU.}}
\begin{tabular}{lcccccc}
\hline
\textbf{Methods}           & \textbf{Segmentor} & \textbf{Generative} & \textbf{Discriminative} & \textbf{RefCOCO}                     & \textbf{RefCOCO+}                   & \textbf{RefCOCOg} \\
\hline
\multicolumn{7}{l}{\textit{Weakly-supervised Method}} \\
TSEG~\cite{strudel_weakly-supervised_2022} & & & & 25.95 & 22.62  & 23.41 \\ 
\hline
 \multicolumn{7}{l}{\textit{Zero-shot Method}} \\
\textsc{Ref-Diff/g} & & \checkmark &  & 21.53 & 22.50 & 27.03 \\
\textsc{Ref-Diff/gs} & \checkmark & \checkmark &  & 29.82 & 30.06 & 30.73 \\
\textsc{Ref-Diff/ds} & \checkmark & & \checkmark & 35.27 & 35.72 & 41.84 \\
\textsc{Ref-Diff} & \checkmark & \checkmark & \checkmark & \textbf{37.21} & \textbf{37.29} & \textbf{44.02} \\
 \hline
\end{tabular}
\label{tab:abl_miou}
\end{table*}

\begin{figure*}
	\centering
	\includegraphics[width=14cm]{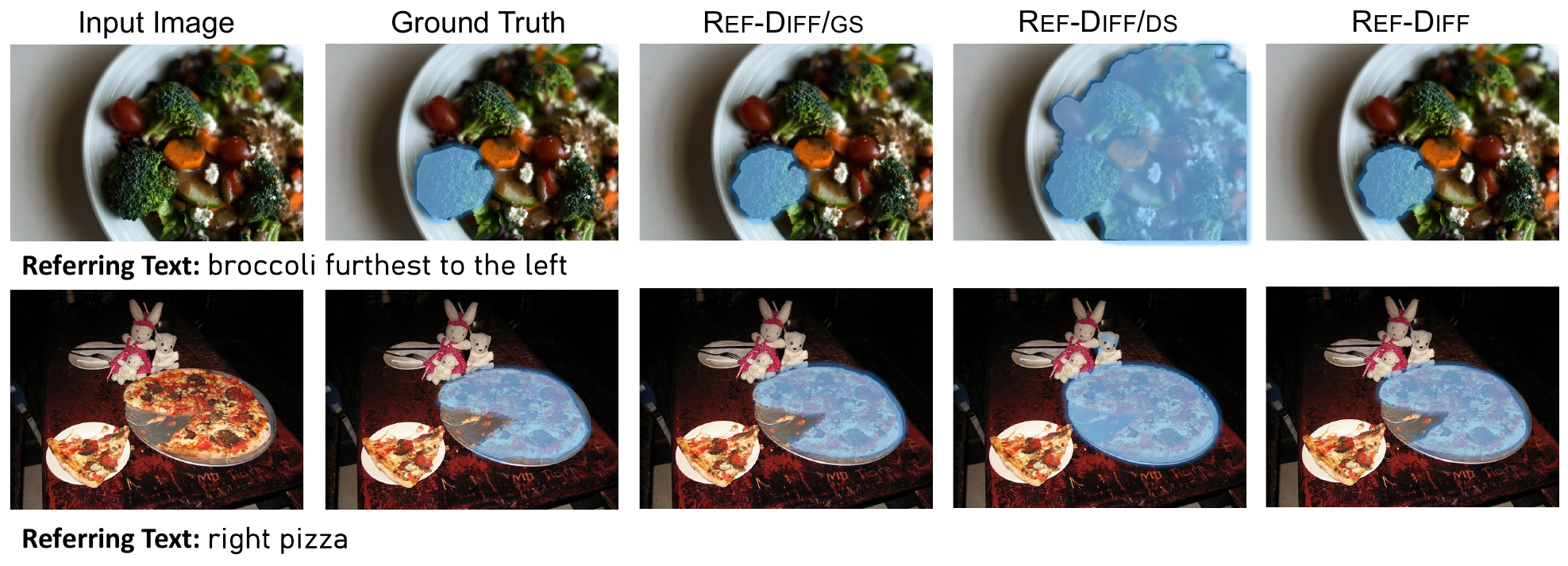}
	\caption{\textbf{Effectiveness of generative model in localization capability.} The discriminative model focuses more on whether the image contains text-related content, which may result in mistakenly selecting larger regions. 
    }
	\label{fig:gen+}
\end{figure*}

\begin{figure*}
	\centering
	\includegraphics[width=14cm]{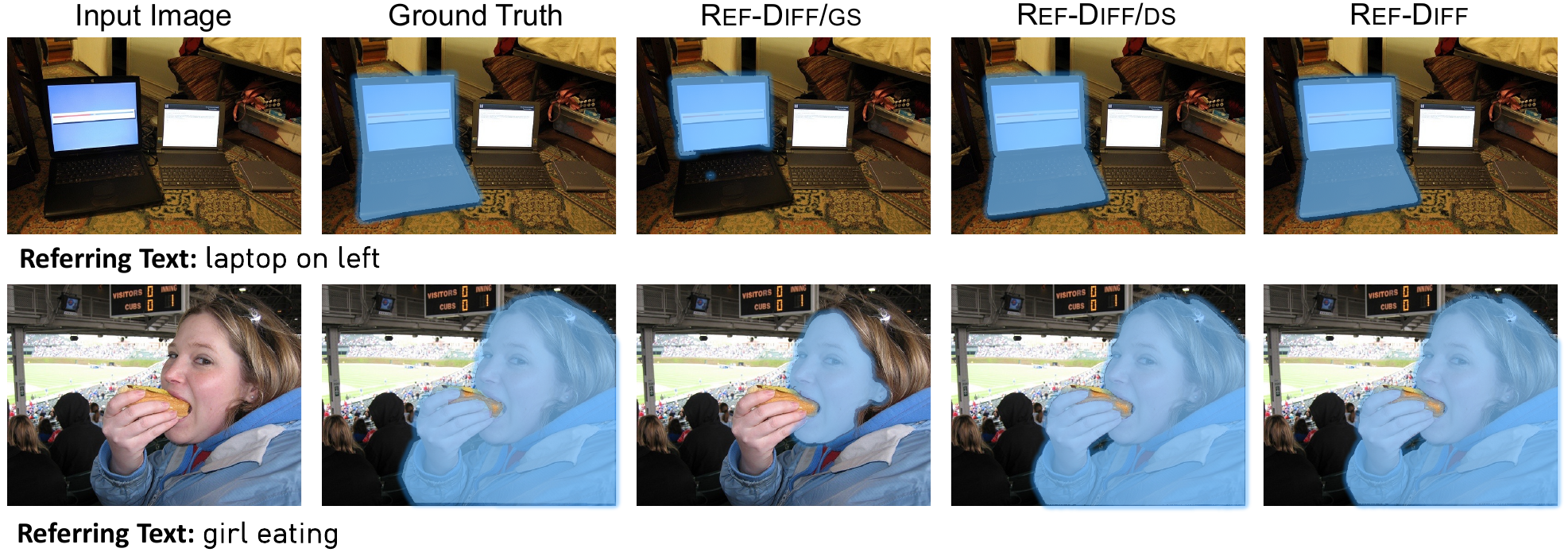}
	\caption{\textbf{Effectiveness of discriminative model.} The generative model exhibits higher sensitivity to salient visual features, which can result in partial segmentation when solely relying on the generative model. By integrating the discriminative model, we can effectively mitigate such errors and achieve more accurate results.}
	\label{fig:clip+}
\end{figure*}

\subsection{Effect of Generative Model}

\paragraph{Segmentation Capability.}

From Figure~\ref{fig:nop} it can be observed that even without the assistance of the segmentor and CLIP, our \textsc{Ref-Diff/g} is able to accurately segment the corresponding content. This is primarily attributed to the attention projection of the generative model onto the relevant visual content (kindly refer to Sec. \ref{sec:attn}). In the second example, despite the presence of two instances of the same object (person) in the image, our \textsc{Ref-Diff/g} is still capable of performing reasonably accurate segmentation. This highlights the significant potential of the generative model, as it exhibits 1) segmentation capabilities comparable to segmentation models and 2) categorization capabilities comparable to discriminative models.

\paragraph{Localization Capability.}

From the first example shown in Figure~\ref{fig:gen+}, it can be observed that the discriminative model fails to accurately locate the leftmost broccoli. Similarly, in the second example, CLIP fails to eliminate the redundant region of the plate. We analyze this issue may arise due to the inherent limitation of the discriminative model, which is trained to identify whether the given text and image are well aligned. So it lacks the capability of localizing objects within the image. Consequently, relying solely on the discriminative model to discern whether a region contains redundant content becomes challenging. However, when we combine the generative and discriminative models, we are able to achieve the best results.

\subsection{Effect of Discriminative Model}

In the first example depicted in Figure~\ref{fig:clip+}, we observe that the generative model incorrectly segments the screen as a separate object due to its prominence as a significant visual feature of a laptop, attracting more attention from the generative model. Similarly, in the second example, the generative model places greater emphasis on the person's face, resulting in incomplete segmentation. However, these errors can be effectively mitigated in our full model \textsc{Ref-Diff}, due to the robust categorization capability of the discriminative model.

\subsection{Attention from Generative Model}
\begin{figure*}
	\centering
	\includegraphics[width=14cm]{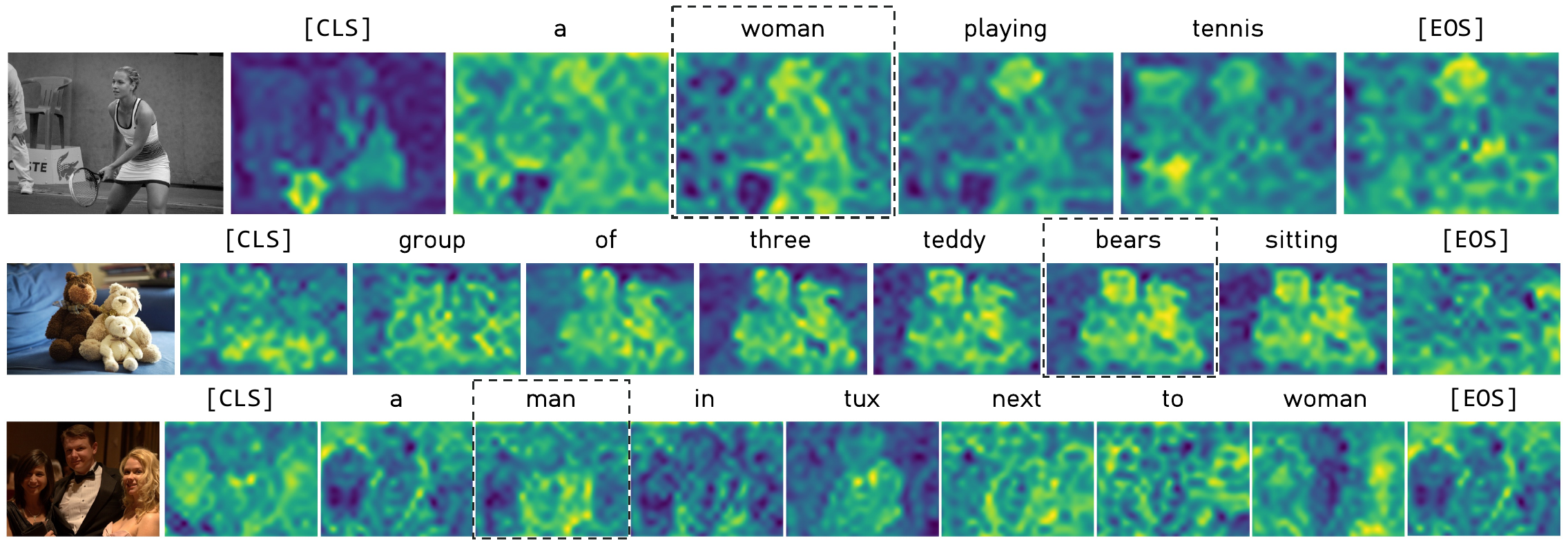}
	\caption{\textbf{Attention from generative model.} Generative model projects attention to different regions of the image based on different tokens, which is the key reason for the effectiveness of Ref-Diff. The dashed box highlights the root token and its corresponding attention map.}
	\label{fig:attn}
\end{figure*}
\label{sec:attn}

To investigate the region that the generative model focuses on, we provide visualizations of sample images along with the attention weights assigned to each token in the generative model, as depicted in Figure~\ref{fig:attn}. We find that: 1) Generative models exhibit contextual understanding capabilities, as they effectively allocate attention to the relevant regions in the image, thereby accomplishing both localization and segmentation tasks. Notably, the attention assigned to the subject token closely aligns with the final segmentation results. This finding elucidates why generative models can perform well in Zero-shot referring segmentation tasks without relying on a separate segmentor or a discriminative model. 2) In contrast to classification models, where the first token often represents the entirety of the text, we observe that the information associated with the first token in our generative model does not capture the complete textual context. However, as discussed earlier, the subject token successfully captures the comprehensive information from the entire text, enabling accurate segmentation results.

\subsection{Case Study}
\label{sec:case}

\begin{figure*}
	\centering
	\includegraphics[width=14cm]{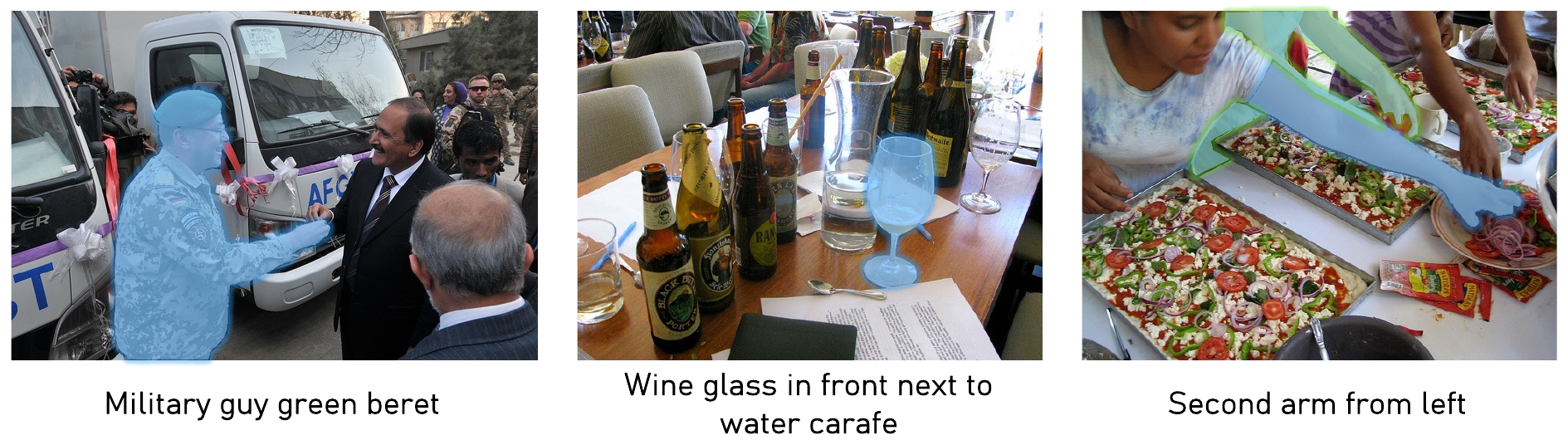}
	\caption{\textbf{Case studies.} Blue indicates the predicted regions. The third case is a failure example, where green denotes ground-truth regions. \textsc{Ref-Diff}  demonstrates its ability to accurately segment objects based on the provided referring texts, even in the presence of complex spatial relationships within the image. }
	\label{fig:case}
 \vspace{-0.2in}
\end{figure*}

We presented three examples of varying difficulty in Figure~\ref{fig:case} to showcase the effectiveness of our \textsc{Ref-Diff}. In the first example, we observed that \textsc{Ref-Diff} demonstrates the capability to accurately identify and segment the correct object within similar objects. In the second example, despite the presence of numerous objects in the image and the complex spatial relationships, \textsc{Ref-Diff} successfully identifies the correct objects through the accurate understanding of the generative and discriminative models. In the final example, we encountered a segmentation failure of \textsc{Ref-Diff} due to the presence of some degree of ambiguity in the referring expression. \textsc{Ref-Diff} incorrectly identifies the leftmost hand as the first arm, resulting in a segmentation error. Enhancing the robustness of \textsc{Ref-Diff} is an area of future work that deserves further investigation.

\section{Broader Impact and Limitations}

Zero-shot Referring Segmentation has broad applications in industrial and real-world domains, such as image editing, robot control, and human-machine interaction. our \textsc{Ref-Diff}, through the combination of generative and discriminative models, has successfully demonstrated the feasibility of Training-free yet high-quality Referring Segmentation in various data-scarce scenarios. This significantly reduces the deployment cost of artificial intelligence in related fields. However, due to the existence of pre-trained modules, the inference stage still incurs high computational overhead. Moreover, Referring Expression texts are sensitive to ambiguity (see Sec.~\ref{sec:case}), which currently results in noticeable segmentation errors. In the future, we will further investigate Zero-shot Referring Segmentation with lower computational costs and higher robustness.

\section{Conclusion}
In this work, we proposed a novel Referring Diffusional segmentor (\textsc{Ref-Diff}) for Zero-shot Referring Image Segmentation, which effectively leverages the fine-grained multi-modal information from generative models. We demonstrated that a generative model alone can achieve comparable performance to existing SOTA weakly-supervised models without requiring a proposal generator. Moreover, by combining both generative and discriminative models, \textsc{Ref-Diff} outperformed these competing methods by a significant margin. Overall, our work presented a simple yet promising direction for zero-shot referring image segmentation by exploiting the potential of generative models, which brought new insights for addressing the challenges of this task.

\bibliographystyle{plain}
\bibliography{neurips_2023}

\end{document}